\documentclass[letterpaper, 10 pt, conference]{ieeeconf}  

\IEEEoverridecommandlockouts
\usepackage{cite}
\usepackage{graphicx}
\usepackage{textcomp}
\usepackage{array}
\usepackage{subcaption}
\usepackage{calc}
\usepackage{placeins}
\usepackage[table]{xcolor}
\usepackage{hyperref}

\title{\LARGE \bf TOG: Training-free Task-oriented Grasp Generation}

\author{
Jiaming Wang$^{*}$, 
Jizhuo Chen$^{*}$,
Diwen Liu$^{*}$, 
Harold Soh%
\thanks{$^{*}$Equal contribution.}%
\thanks{Emails: jiaming@comp.nus.edu.sg, \{e0774920, e0905370\}@u.nus.edu, harold@comp.nus.edu.sg.}
}

\begin{document}

\maketitle
\thispagestyle{empty}
\pagestyle{empty}

\begin{abstract}
We present a \textbf{t}raining-free pipeline for task-\textbf{o}riented \textbf{g}rasp generation (TOG) that integrates pre-trained grasp generation models with vision-language models (VLMs). Beyond focusing on stability, our method incorporates task-specific requirements by leveraging the semantic reasoning capabilities of VLMs. We evaluate different strategies for integrating the VLM with a pre-trained grasp generation model, demonstrating through extensive empirical studies that this simple yet effective approach performs robustly in both simulated environments and real-world robotic experiments. Our results highlight the potential of VLMs to enhance task-oriented manipulation, offering valuable insights for future research in robotic grasping and human-robot interaction. Codes can be found at: https://github.com/StevenLiudw/TOG
\end{abstract}


\section{Introduction}

Several grasp generation methods have been introduced in recent research \cite{fang2023anygrasp, mousavian20196, mahler2017dex, sundermeyer2021contact}, with the majority focusing on achieving stable, collision-free grasps. While stability and collision avoidance are critical, human grasping behavior is inherently task-driven. For humans, grasping is not just about securing an object safely but about fulfilling a specific goal. For instance, when picking up a cup to drink, a person typically grasps it by the handle, even though other stable, collision-free grasps (like holding the cup’s body) are possible. This task-oriented approach highlights that grasp selection is based on the intended use of the object.

To enable service robots to effectively perform a wide variety of tasks in real-world environments, it is crucial that they learn to predict task-oriented grasps for a diverse range of objects. For example, a robot may need to grasp a tool differently depending on whether it is going to use it for cutting, hammering, or passing it to a person. Such task-oriented grasp prediction is essential for improving a robot's functionality in practical applications~\cite{murali2021same}.

Training a task-oriented grasp generation policy directly would require an extensive dataset to account for the vast diversity of objects, which presents significant challenges in terms of data collection and cost. Additionally, there are concerns regarding whether such a trained policy would generalize effectively to unseen objects and tasks. Achieving this level of generalization would demand a semantic understanding of both the task and the properties of unfamiliar objects, which is difficult to achieve through training on datasets that consist primarily of grasp annotations.

\begin{figure}[t]
    \centering
    \includegraphics[width=0.9\linewidth]{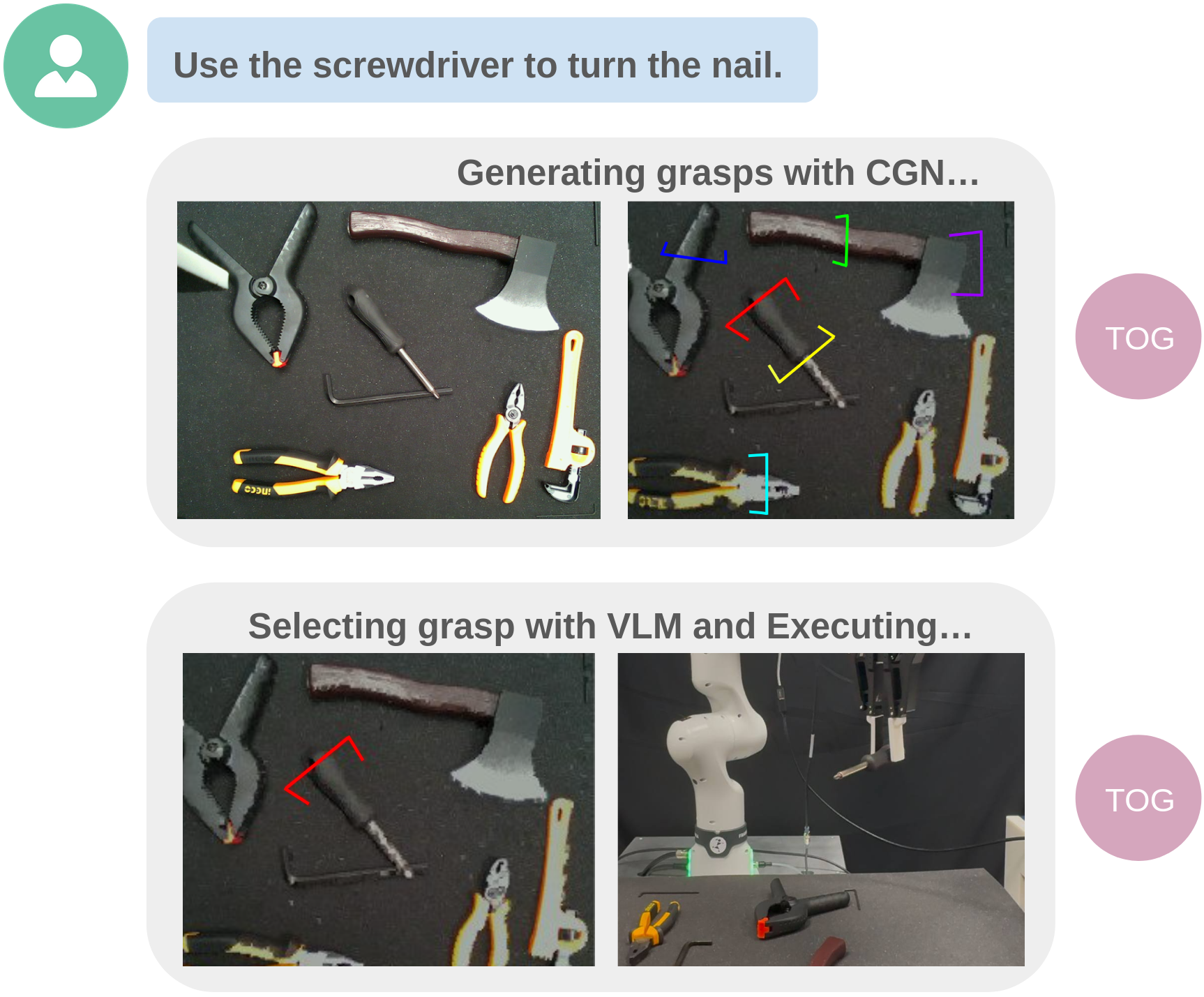} 
    \caption{Given a specific task, TOG can generate feasible grasps according to the task specifications by leveraging the combined capabilities of  (CGN) and a Vision-Language Model (VLM).}
    \label{fig:intro}
\end{figure}

In contrast, vision language models (VLMs) \cite{yang2023dawn, li2023m}, which are trained on massive, internet-scale datasets, have demonstrated remarkable generalization capabilities in understanding both natural language and images. This raises an intriguing question: \textit{can VLMs be leveraged to aid in task-oriented grasp generation?} Several studies have explored this potential \cite{barad2023graspldm, taunyazov2023grace}. However, these approaches often require additional training steps to generate task-oriented grasps (see Section \ref{review}) and tend to be tightly integrated with a particular grasp generation model. This rigidity limits flexibility and adaptability, making it challenging to incorporate more advanced grasp generation models as they emerge.

In this paper, we describe steps towards a \textbf{training-free pipeline} for \textbf{task-oriented grasp generation} (\textbf{TOG}) by integrating a pre-trained grasp generation model with a VLM. Our key finding is that for more capable VLMs, allowing them to freely specify grasp points and matching them to generated grasps yields the best performance. In contrast, for less powerful models, a more constrained approach---where the VLM selects from pre-filtered grasps---leads to better results.

\begin{figure*}[tb]  
  \centering
  \includegraphics[width=\textwidth]{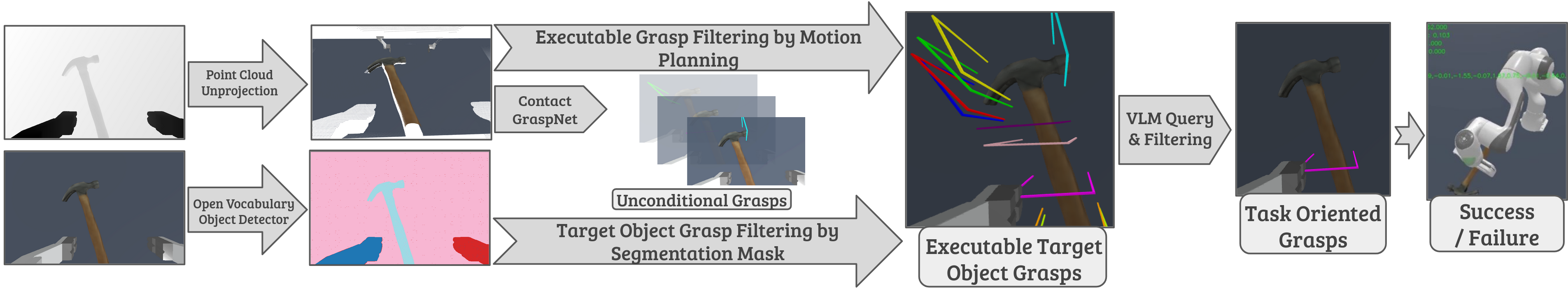}
  \caption{Overview of the system. The depth image is unprojected into a point cloud, which is processed by a grasp generation model (e.g., ) to produce a set of unconditional grasps. The top K grasps are selected based on confidence and further refined by a motion planner to ensure trajectory feasibility. These filtered grasps are then evaluated using a vision-language model (VLM) to select the best grasp for the task.}
  \label{fig:overview}
\end{figure*}

We evaluated and compared different approaches to integrating grasp generation models with VLMs in both a simulation environment \cite{gu2023maniskill2} and real robot settings. Our experiments demonstrate that this simple yet effective strategy is highly robust, achieving an average 70\% success rate on real robots. This study provides insights into how pre-trained models and VLMs can be seamlessly integrated to enhance task-oriented grasping.

To summarize, the main contributions of this work are as:
\begin{itemize}
    \item A simple, flexible, and training-free pipeline for task-oriented grasp generation;
    \item An analysis of different integration approaches, including failure modes and insights to guide future research. 
\end{itemize}


\section{Literature Review} \label{review} \subsection{Non-task-oriented Grasp Generation} 
There exists a substantial body of research on general grasp generation methods. For instance, \cite{mousavian20196} trained a variational autoencoder (VAE) to sample grasps in cluttered scenes, while \cite{urain2023se} utilized diffusion models to capture the multimodal distribution of grasps for a given object. \cite{sundermeyer2021contact} proposed a method that samples dense grasp poses for each possible contact point on an object. Additionally, \cite{fang2023anygrasp} developed a robust grasp generation model trained on noisy real-world data. These approaches successfully generate grasps that avoid collisions and generalize well across a wide variety of objects. However, they do not account for the specific tasks associated with manipulating the object.

\subsection{Task-oriented Grasp Generation} 
In contrast, task-oriented grasp generation focuses on predicting grasps that are optimized for a specific task. Recent work~\cite{murali2021same, jang2017end} introduced data-driven approaches, where large datasets were collected to train neural networks capable of predicting task-oriented grasps. More recently, several studies have explored the integration of large language models (LLMs) and vision language models (VLMs) to enhance task-oriented grasp generation. For example, \cite{tang2023task, li2024semgrasp} leveraged language information alone, while \cite{chang2024text2grasp} required the training of additional grasp generation models. \cite{taunyazov2023grace} employed differentiable classifiers to guide the grasp sampling process; however, while this approach is training-free, it is not directly applicable to more general scenarios where gradient information is not available.

\subsection{Foundation Models in Robotics} 
Building on the success of foundation models, there is growing interest in the use of LLMs and VLMs in various robotic applications. With their large capacities, these models can be trained or fine-tuned to directly predict a sequence of actions given a task description in natural language and RGB observations \cite{liu2024moka, team2024octo, brohan2023rt}. However, training such large models for open-world scenarios typically requires extensive data and supervision.

An alternative approach involves leveraging pre-trained VLMs in combination with lower-level skills to perform open-vocabulary manipulation tasks \cite{weerakoon2024behav, nasiriany2024pivot} or navigation tasks \cite{liu2024moka, wu2024helpful}. Nevertheless, recent studies \cite{ramakrishnan2024does} suggest that foundation models still struggle with spatial reasoning in general. In this work, we aim to investigate whether VLMs can effectively comprehend and reason about the spatial and semantic information related to grasping tasks. Unlike MOKA \cite{liu2024moka}, which assumes a fixed tabletop setup, our approach focuses on task-oriented grasp generation using VLMs. This approach is not constrained to a specific environment, offering a flexible framework that integrates various grasp generation models with VLMs to accommodate a wider range of scenarios. This adaptability enables a more versatile approach to task-oriented manipulations.

\begin{figure}[tb]  
  \centering
  \includegraphics[width=\columnwidth]{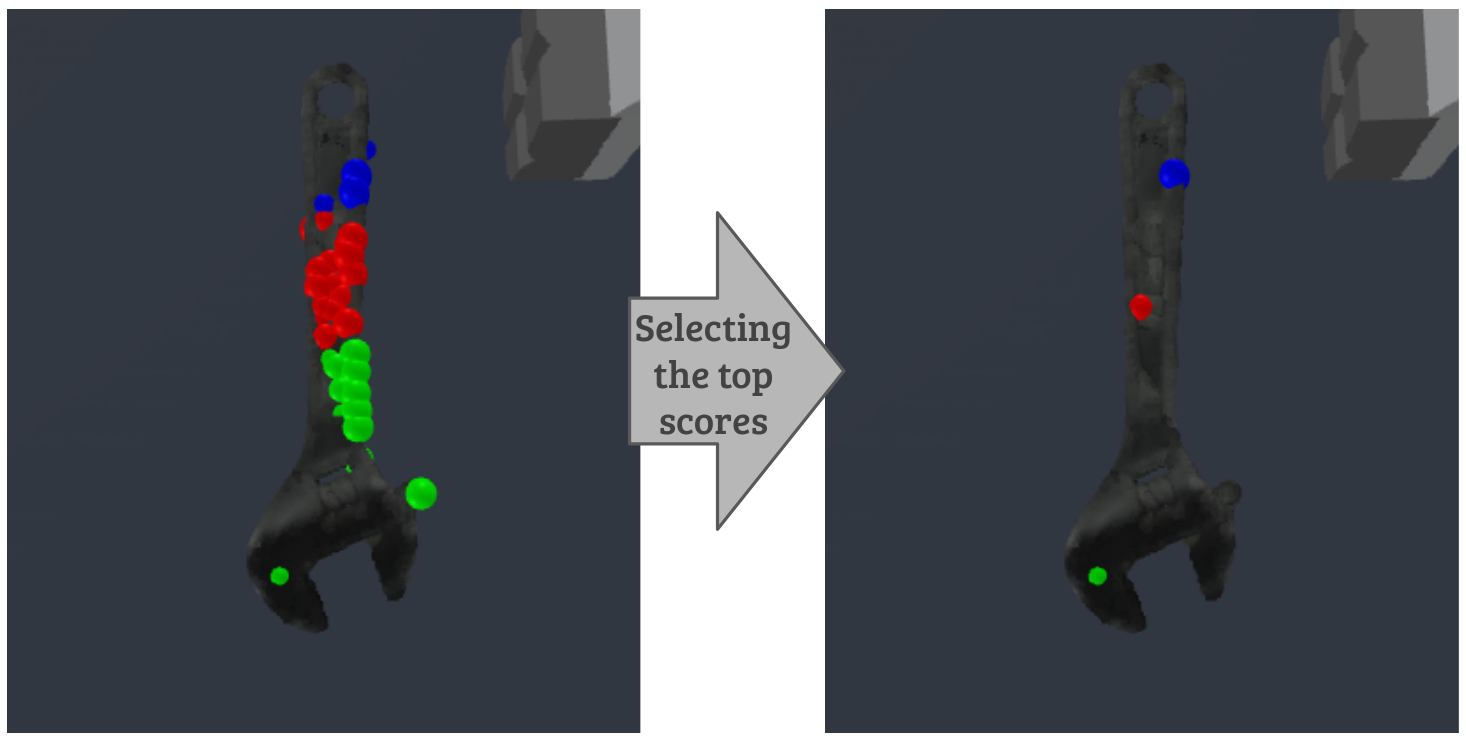}
  \caption{Illustration of using K-means clustering and scores to select diverse grasps while taking into account the quality of each grasp.}
  \label{fig:kmeans}
\end{figure}

\section{Task-oriented Grasp Generation}
\label{task-oriented-grasp-generation}

\subsection{Task Definition}
We address the problem of generating task-oriented grasps for robotic manipulation, given the point cloud of an object and task-oriented constraints expressed in natural language. Our objective is to estimate the grasp distribution \( P(G^* \mid X, T) \), where \( X \) represents the RGB and depth image inputs, \( T \) denotes the task-oriented constraints, and \( G^* \) refers to the set of successful grasps that satisfy both stability and task-oriented requirements. As in prior work \cite{sundermeyer2021contact, fang2023anygrasp}, each grasp \( g \in G^* \) is defined as a grasp pose \( (R, t) \in SE(3) \) for a parallel-jaw gripper.

For the task-oriented constraints \( T \), we consider natural language task descriptions that provide guidance on how the object should be grasped. For instance, given a pair of scissors and the task description ``passing them to someone,'' the robot should grasp the blade area; whereas for the task ``use it,'' the robot should grasp the handle.

\subsection{System Overview}
\label{overview}

Our proposed approach generates the optimal task-oriented grasp in two stages: 
(1) sampling a set of task-agnostic grasps from a pretrained grasp generation model, and 
(2) filtering and selecting the grasp that best satisfies the task constraints. 
This decomposition enables us to leverage existing work on stable grasp generation while integrating task-specific constraints to ensure the selected grasp is well-suited for the given task. 
An overview of our proposed system is illustrated in Fig.~\ref{fig:overview}.

Given an RGB and depth image as input, our pipeline begins by processing the depth image to generate a point cloud through unprojection. This point cloud is then input into a pre-trained grasp generation model to produce a set of candidate grasps \( \{G_i\}_{i=1}^N \). For our experiments, we utilized  \cite{sundermeyer2021contact} due to its robustness. However, our pipeline is flexible and can accommodate other, potentially more advanced, grasp generation models.

Generated grasps are passed to a motion planner, which filters out any grasps that are not feasible due to reachability or collision constraints. This step is performed prior to querying the VLM to reduce computational overhead.

\begin{table}[t!]
\setlength{\belowcaptionskip}{-12pt}
\centering
\caption{VLM querying methods.}
\label{tab:query-methods}
\small
\begin{tabular}{|m{0.44\linewidth}|m{0.44\linewidth}|}
\hline
\multicolumn{1}{|c|}{\textbf{Image Prompt}} & \multicolumn{1}{c|}{\textbf{Text Prompt}} \\
\hline
\multicolumn{2}{|c|}{Contact Points in a Single Image (CPSI)} \\
\hline
\begin{center}
\includegraphics[width=0.9\linewidth]{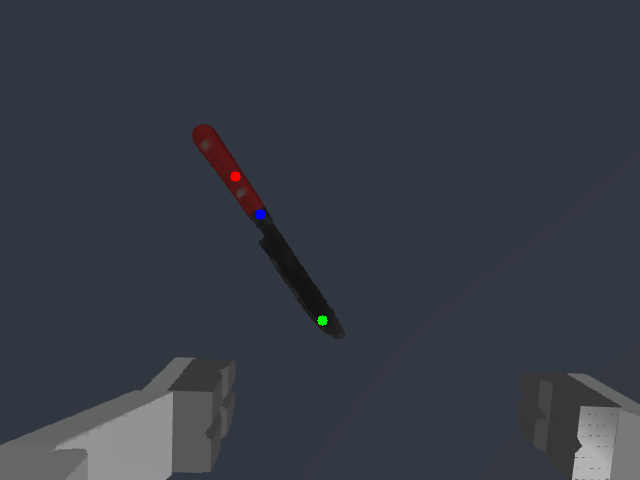}
\end{center}
& Given \textbf{an image} showing the object with different grasp candidates \textbf{highlighted in red, green, and blue dots}, your task is to analyze these grasp candidates \textit{based on the task description} and choose the index of the grasp that would \textit{best accomplish the task.} \\
\hline
\multicolumn{2}{|c|}{Contact Points in Multiple Images (CPMI)} \\
\hline
\begin{center}
\includegraphics[width=0.9\linewidth]{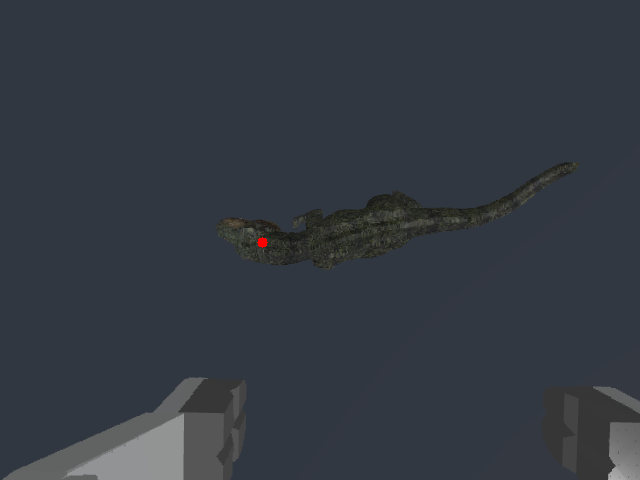}
\end{center}
& Given \textbf{multiple images}, each showing the object with a different grasp candidate \textbf{with a red dot}, and a task description, your task is to analyze the grasp candidates in \textbf{each image} \textit{based on the task description} and choose the index of the grasp that would \textit{best accomplish the task.}\\
\hline
\multicolumn{2}{|c|}{Grasps in a Single Image (GSI)} \\
\hline
\begin{center}
\includegraphics[width=0.9\linewidth]{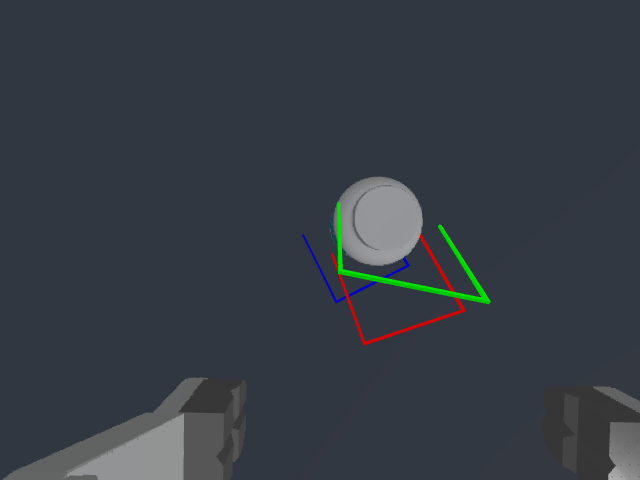}
\end{center}
& Given \textbf{an image} showing the object with different grasp candidates \textbf{highlighted in red, green, and blue}, your task is to analyze these grasp candidates \textit{based on the task description} and choose the index of the grasp that would \textit{best accomplish the task.}\\
\hline
\multicolumn{2}{|c|}{Grasps in Multiple Images (GMI)} \\
\hline
\begin{center}
\includegraphics[width=0.9\linewidth]{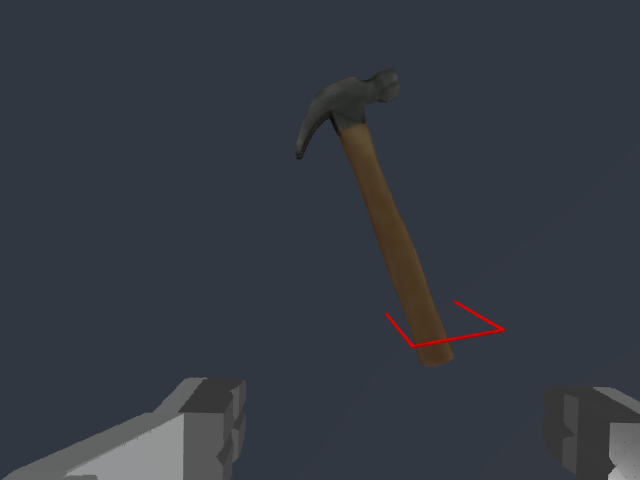}
\end{center}
& Given \textbf{multiple images}, each showing the object with a different grasp candidate \textbf{highlighted in red}, and a task description, your task is to analyze the grasp candidates in \textbf{each image} \textit{based on the task description} and choose the index of the grasp that would \textit{best accomplish the task.}\\
\hline
\multicolumn{2}{|c|}{Contact Point Generation (CPG)} \\
\hline
\begin{center}
\includegraphics[width=0.9\linewidth]{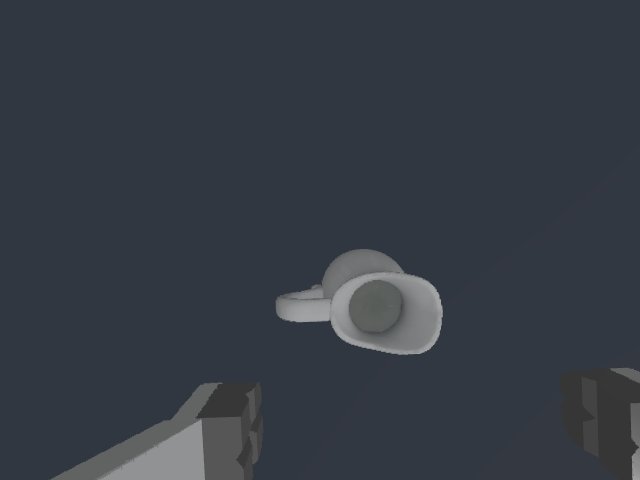}
\end{center}
& Given an image of the scene and a task description, you should \textbf{provide a point} on the image where the robot should grasp the object to \textit{best accomplish the task.}\\
\hline
\end{tabular}
\end{table}

One limitation of the proposed pipeline is that the pre-filtered top-\(k\) grasp candidates (based on the confidence scores from the grasp generation model) often lack diversity. These candidates typically belong to the same cluster of grasps. For instance, in the case of a screwdriver, most top-rated grasps focus on the midpoint of the handle (Fig. \ref{fig:failure_modes} (a)). While these grasps yield a high grasp success rate, they are less useful for task-oriented grasp generation, which requires a diverse set of grasps that can then be ranked by the VLM.

To enhance grasp diversity, we generate N valid grasps and pre-filter them using the motion planner, as described in Section \ref{overview}. We then apply \(k\)-means clustering to the contact points (i.e., the center of the gripper) in Euclidean space. The effect of different \(k\) values is analyzed in Section \ref{sec:compare-k}. From each cluster, we select the grasp with the highest confidence score predicted by the grasp generation model. This approach ensures that the final set of candidate grasps is both diverse and of high quality (Fig. \ref{fig:kmeans}).

It is worth noting that more sophisticated clustering methods could also be employed. However, in this work, we use the simpler strategy outlined above, as it empirically yields satisfactory results.

\subsection{Task-Oriented Grasp Filtering by VLM}
\label{vlm-filter}

After generating K grasp candidates, we query the VLM to select the grasp most suitable for the specified task. Queries are constructed by combining the grasp candidates with the task description and submitting them to the VLM. The VLM's response is then used to select the best grasp for the task.

We employ five distinct strategies to query the VLM, each differing in how candidate grasps are visualized and presented.

\begin{enumerate}
    \item \textbf{Grasps in a Single Image (GSI):} All candidate grasps are visualized in a single image, with each grasp represented as a U-shaped wireframe at its corresponding pose, projected onto the observed RGB image. This composite visualization allows the VLM to compare all grasps directly within a single scene.

    \item \textbf{Contact Points in a Single Image (CPSI):} Candidate grasps are visualized as colored dots representing their contact points in a single image. The VLM selects the optimal grasp based on the task description, enabling direct comparison of contact points within a single view.

    \item \textbf{Grasps in Multiple Images (GMI):} Each candidate grasp is displayed in a separate image, where the grasp pose is represented by a gripper wireframe projected onto an individual RGB observation. The VLM then selects the image containing the grasp best suited for the task.

    \item \textbf{Contact Points in Multiple Images (CPMI):} Each candidate grasp is represented by a colored dot indicating its contact point, displayed in a separate image. This simplified visualization reduces visual complexity for the VLM, allowing it to focus solely on contact point selection.

    \item \textbf{Contact Point Generation (CPG):} Instead of providing predefined grasp candidates, we directly query the VLM to suggest an optimal grasping point in the image by specifying pixel coordinates \((u, v)\). To ensure the selected point can be unprojected into 3D space, we explicitly instruct the VLM to identify a point on the target object. We then select the 3D point \(P\) whose projection onto the image plane is closest to the queried \((u, v)\) coordinate. Finally, we choose the grasp whose contact point is nearest to the 3D point \(P\).
\end{enumerate}

An illustration of these query methods is provided in Table~\ref{tab:query-methods}.

\
\section{Simulation Experiment}
\label{experiment}

We conducted experiments in the simulator to evaluate the TOG pipeline using various VLMs and querying strategies. Our goal was to assess the effectiveness of integrating pre-trained grasp generation models with VLMs to select grasps that satisfy specific task requirements.

\subsection{Experimental Setup}

Our experiments were conducted in the ManiSkill simulator~\cite{gu2023maniskill2} using the Franka Emika Panda robotic arm equipped with a parallel-jaw gripper. We selected 29 objects from the YCB~\cite{calli2015ycb} and Google Scanned Objects (GSO)~\cite{borja2014googlescan} datasets, ensuring diversity in shape, size, and functionality.  

To evaluate task-oriented grasp generation, we defined three task categories: \emph{Grasping at a Specific Position, Tool Use, and Object Handover}. Each task required the robot to grasp the object at a predefined position based on its intended function (see Table~\ref{table:task_description}).  

\begin{table}
\renewcommand{\arraystretch}{1.3} 
\setlength{\belowcaptionskip}{-4pt}
\centering
\caption{Three categories of tasks used in simulation experiment}
\label{table:task_description}
\begin{tabular}{|p{0.22\linewidth}|p{0.27\linewidth}|p{0.35\linewidth}|} 
\hline
\textbf{Task Type} & \textbf{Example} & \textbf{Success Condition} \\
\hline
\textbf{Grasping at a Specific Position} & 
Grip the toy by its tail. & 
The toy is grasped precisely at the specified position. \\
\hline
\textbf{Tool Usage} & 
Hold the screwdriver by the handle for ease of use. &  
The tool is held in a way that facilitates its intended function. \\
\hline
\textbf{Object Handover} & 
Pass me the screwdriver. &  
Hold the screwdriver by the shaft, leaving the handle free for easy transfer. \\
\hline
\end{tabular}
\end{table}

For evaluating task compliance, we initially attempted an automated assessment using Qwen-72B~\cite{Qwen2VL}. To verify its reliability, we manually reviewed a subset of results and compared them with Qwen-72B’s judgments. The model’s precision—defined as the agreement between its predictions and human evaluations—was 0.717. Given this suboptimal accuracy, we opted for manual evaluation following the predefined protocols in Table~\ref{table:task_description}.  

We employed  (CGN) to generate candidate grasps for each object. CGN outputs grasps with associated quality scores, which were filtered using a motion planner to remove kinematically infeasible or unreachable grasps. We then followed the methodology described in Section~\ref{task-oriented-grasp-generation} to generate a diverse set of grasp candidates. For the $k$-means algorithm, we set $k=3$, as it is robust across different VLMs.

Our experiments compared three Vision-Language Models (VLMs):

\begin{itemize}
    \item \textbf{Qwen-7B}~\cite{Qwen2VL}: An open-source VLM known for its strong language understanding and visual reasoning capabilities.
    \item \textbf{Molmo}~\cite{deitke2024molmo}: A VLM designed for robotic applications, particularly point-based interactions, making it well-suited for the CPSI strategy.
    \item \textbf{Gemini-2.0-Pro}~\cite{pichai2024gemini}: A proprietary VLM demonstrating strong performance across diverse tasks.
\end{itemize}

We compared our proposed system against a baseline that selects the grasp with the highest confidence score from the grasp generation model.

\begin{table*}[htp!]
\centering
\caption{Experiment results of the proposed pipeline using different VLM query strategies for each VLM.}
\begin{tabular}{|l|c|c|c|}
\hline
\textbf{Method (VLM)} & \textbf{Grasp Success Rate} & \textbf{Task Compliance Rate} & \textbf{Combined Success Rate} \\ \hline
\multicolumn{4}{|l|}{\textbf{Qwen-7B}} \\ \hline
GSI  & 0.754 & 0.561 & 0.439 \\ \hline
CPSI & 0.702 & 0.491 & 0.386 \\ \hline
GMI  & 0.754 & 0.526 & 0.386 \\ \hline
CPMI & 0.702 & 0.649 & 0.351 \\ \hline
CPG  & 0.737 & 0.614 & 0.544 \\ \hline
\multicolumn{4}{|l|}{\textbf{Molmo}} \\ \hline
GSI  & 0.649 & 0.544 & 0.421 \\ \hline
CPSI & 0.825 & 0.614 & 0.526 \\ \hline
GMI  & 0.737 & 0.579 & 0.351 \\ \hline
CPMI & 0.702 & 0.544 & 0.474 \\ \hline
CPG  & 0.754 & 0.649 & 0.456 \\ \hline
\multicolumn{4}{|l|}{\textbf{Gemini-2.0-pro-exp-02-05}} \\ \hline
GSI  & 0.704 & 0.685 & 0.463 \\ \hline
CPSI & 0.741 & 0.648 & 0.481 \\ \hline
GMI  & 0.611 & 0.778 & 0.481 \\ \hline
CPMI & 0.574 & 0.778 & 0.463 \\ \hline
CPG  & \textbf{0.796} & \textbf{0.778} & \textbf{0.611} \\ \hline
\textbf{Baseline} & 0.754 & 0.298 & 0.175 \\ \hline
\end{tabular}
\label{table:performance}
\end{table*}

\begin{figure*}[h]
    \centering
    \includegraphics[width=0.32\textwidth]{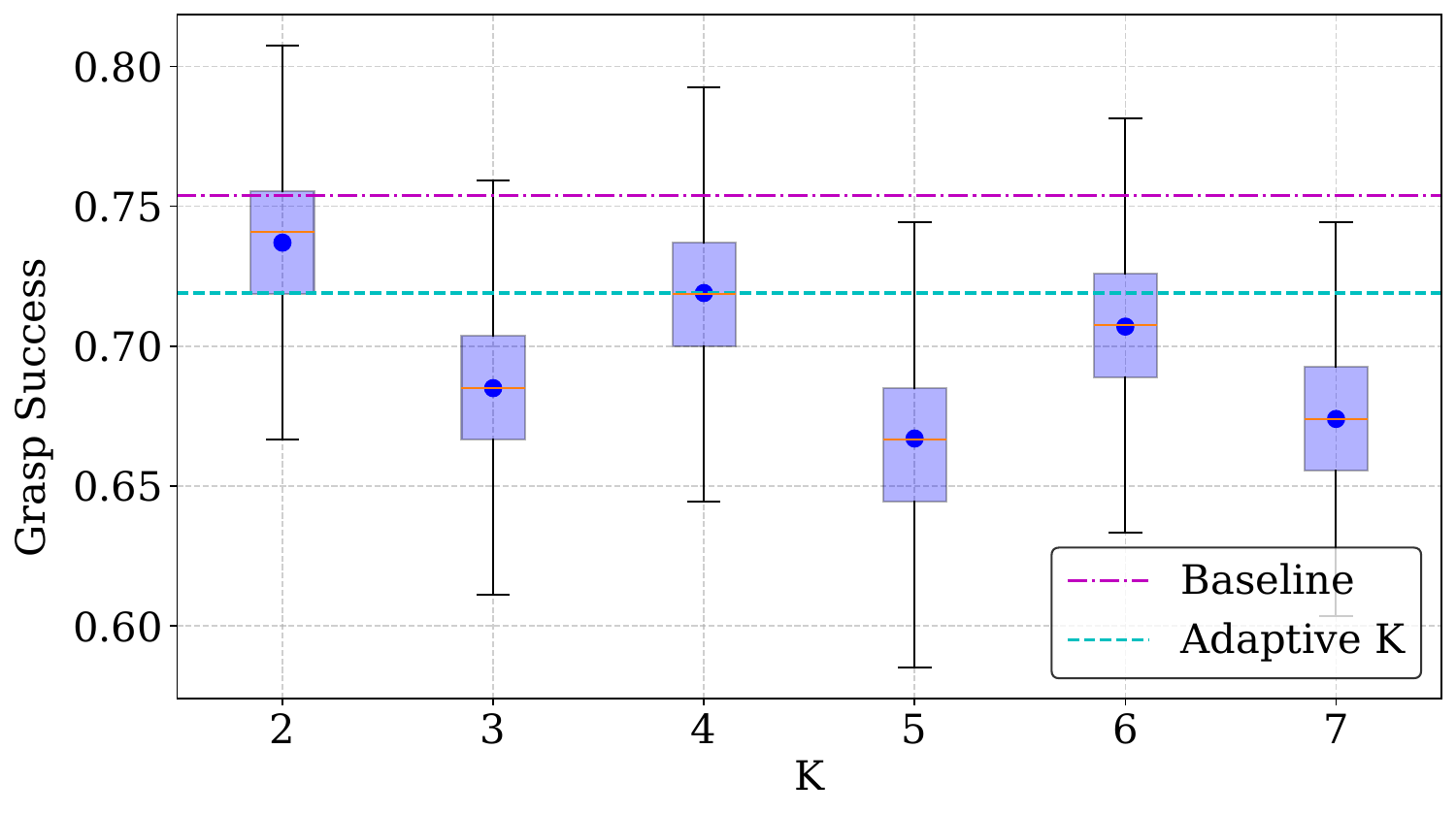}
    \includegraphics[width=0.32\textwidth]{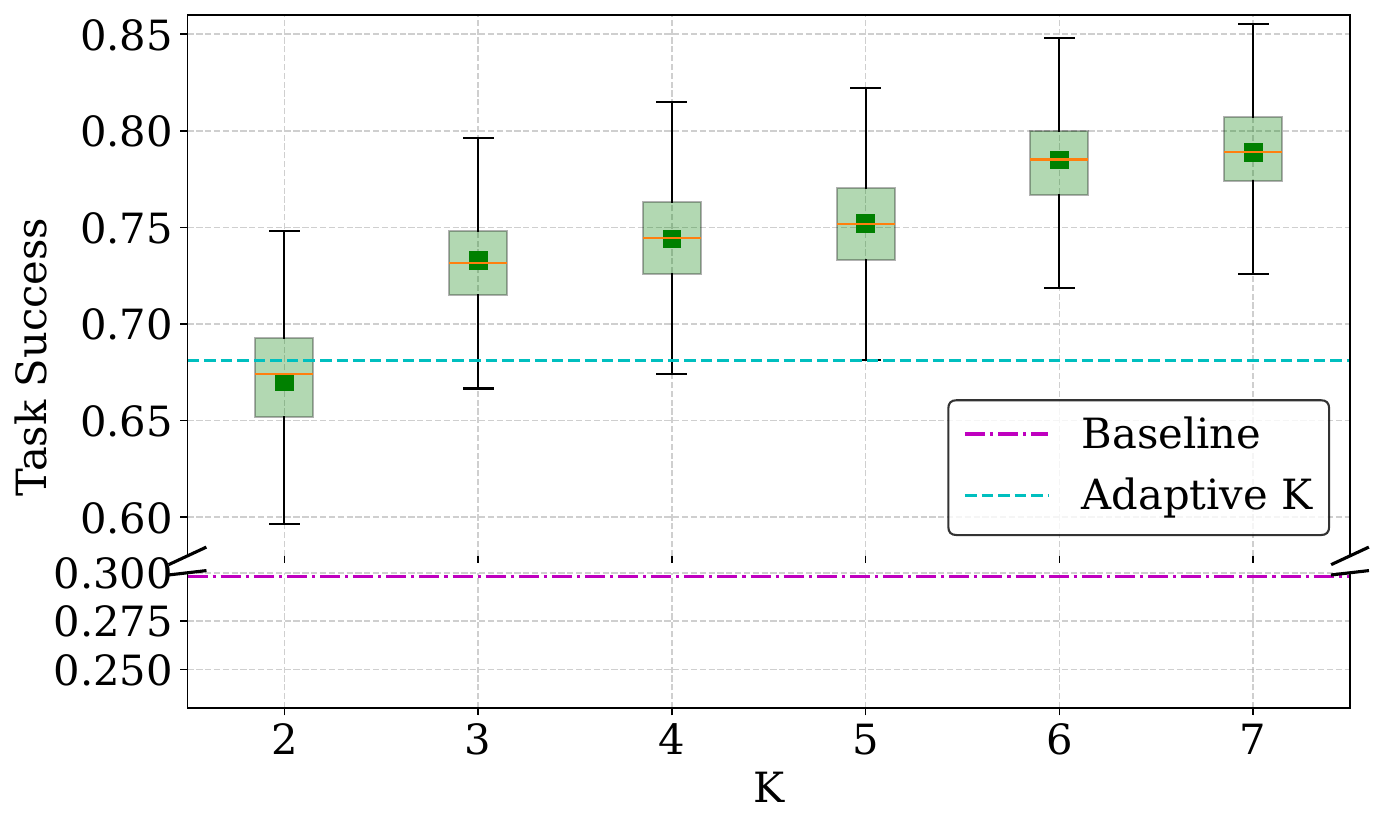}
    \includegraphics[width=0.32\textwidth]{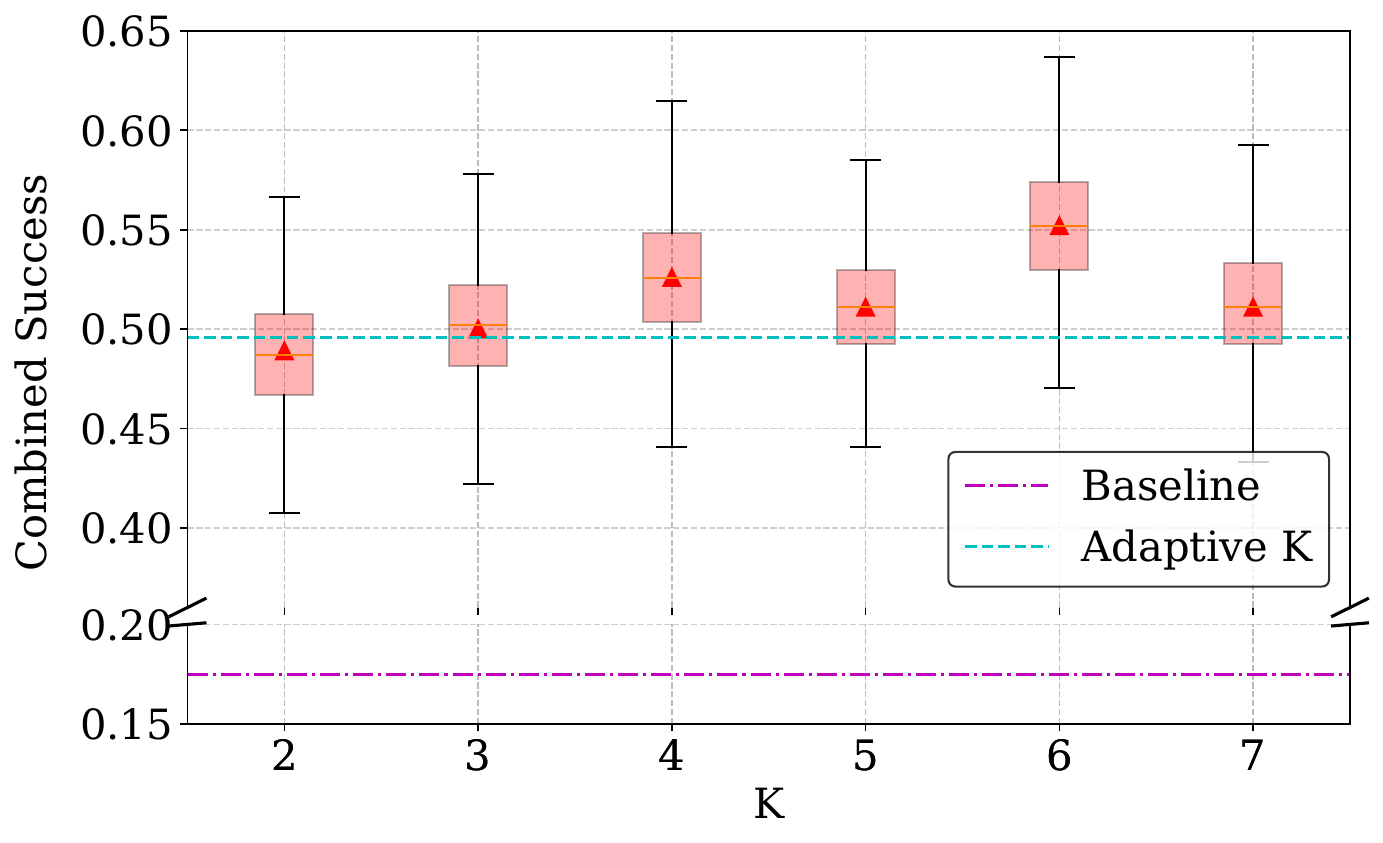}
    \caption{Comparison of grasp success, task success, and combined success across different evaluation metrics.}
    \label{fig:ablation-k}
\end{figure*}

\subsection{Evaluation Metrics}

The performance of our approach was evaluated using the following metrics:

\begin{itemize}
    \item \textbf{Grasp Success Rate:} The percentage of trials in which the robot successfully grasps the object and lifts it above a specified height without dropping it.
    \item \textbf{Task Compliance Rate:} The percentage of trials in which the VLM selects a grasp that aligns with the specific task requirements.
    \item \textbf{Combined Success Rate:} The percentage of trials in which the executed grasp both lifts the object successfully and meets the task-oriented requirements.
\end{itemize}

\subsection{Observations and Analysis}

Table~\ref{table:performance} summarizes our experimental results. Notably, the TOG system consistently outperforms the baseline system by a significant margin, demonstrating the effectiveness of the proposed pipeline for task-oriented grasp generation. 

\begin{figure}[h]
    \centering
    \begin{tabular}{cc}
        \includegraphics[width=0.40\linewidth, viewport=0 0 420 315 , clip]{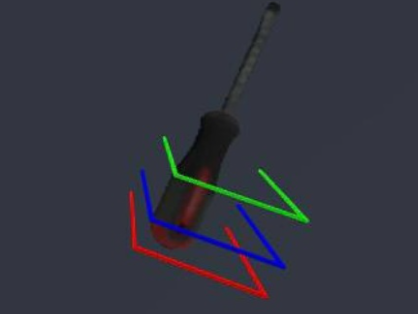} &
        \includegraphics[width=0.40\linewidth, viewport=0 0 420 315, clip]{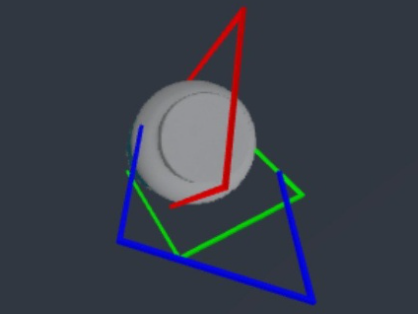} \\
        (a) & (b) \\
        \includegraphics[width=0.40\linewidth, viewport=0 0 420 315, clip]{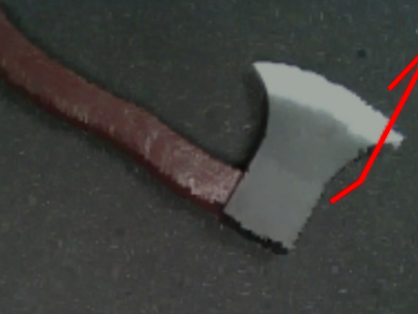} &
        \includegraphics[width=0.40\linewidth, viewport=0 0 420 315, clip]{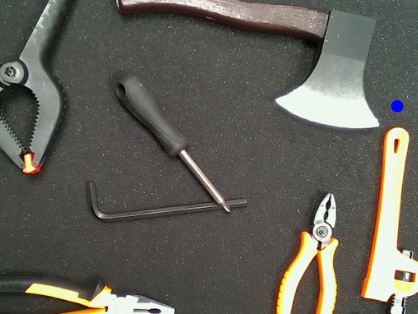} \\
        (c) & (d) \\
    \end{tabular}
    \caption{Common failure cases: (a) Insufficient grasp diversity due to clustering. (b) Ambiguous query image makes grasp identification difficult. (c) VLM selects an unstable grasp. (d) VLM assigns an incorrect contact point (the blue dot)}
    \label{fig:failure_modes}
\end{figure}

{\bf Is TOG effective for generating task-oriented grasps?}  
Among the evaluated strategies, the Contact Point Generation (CPG) method achieves the best performance when combined with Gemini-2.0-Pro, as shown in Table~\ref{table:performance}. However, CPG is less effective when used with weaker VLMs compared to other strategies. This result suggests that when leveraging a more powerful VLM, directly querying it to select the optimal grasp point based on the task description provides an accurate heuristic for selecting grasps generated by the CPG model. In contrast, when using a less powerful VLM, strategies that \textbf{restrict} the VLM to choose from a predefined set of clustered grasps perform better. Further analysis reveals that for weaker VLMs, the generated contact points are often unstable and sometimes misaligned with the object, leading to failures in grasp selection (see Figure~\ref{fig:failure_modes} (c) and (d)). A key insight from our findings is that when using a more capable VLM, providing it with greater "freedom" improves performance, whereas more restrictive approaches yield better results for less powerful models.

\textbf{Does using a VLM affect grasp quality?}  
We observe that the grasp success rate for methods incorporating the VLM is comparable to that of the baseline, indicating that integrating the VLM does not degrade grasp quality. Interestingly, the grasp success rate even increases when using a more powerful VLM in conjunction with the CPG method. We hypothesize that the VLM may implicitly favor more stable grasps, contributing to this improvement.

\textbf{Which query strategy performs best?}  
We observe that methods utilizing multiple images (GMI and CPMI) outperform those using a single image (GSI and CPSI), particularly when paired with Gemini-2.0-Pro. However, for less powerful VLMs, an increased number of images may introduce confusion, leading to a higher failure rate. Regarding query representation, we find that both strategies — visualizing the gripper shape (GSI and GMI) or the contact point (CPSI and CPMI) — perform comparably overall.


\textbf{How to select the number of clusters \(K\)?}  
\label{sec:compare-k}  
We further investigate the effect of different \(K\) values in our experiments. Figure~\ref{fig:ablation-k} presents the results using Gemini-2.0-Pro. We compute the average score across all strategies that utilize clustering (GMI, CPMI, GSI, and GMI) and plot it against different \(K\) values used in the \(k\)-means clustering algorithm. Notably, the task compliance rate increases as \(K\) increases, whereas the grasp success rate decreases. This aligns with our observation that a higher \(K\) allows the VLM greater freedom to select a grasp that best matches the task description. However, this increased freedom comes at a cost: greater diversity in clusters leads to a decline in grasp quality. High-quality grasps are typically supported by more neighboring grasps, which tend to form larger clusters. As a result, increasing \(K\) leads to a trade-off between grasp diversity and grasp success rate. 

In general, we find that selecting the optimal \(K\) by maximizing the average silhouette score provides a good balance between grasp success rate and task compliance rate, as illustrated in Figure~\ref{fig:ablation-k}.

\subsection{Failure Modes}
\label{sec:failures}
Although our proposed system demonstrates promising results in task-oriented grasp generation in a training-free manner, several limitations were observed during the experiments:

\textbf{Clustering Method Does Not Provide Sufficiently Diverse Clusters:} The clustering method used in our approach sometimes fails to generate clusters that are diverse enough. As shown in Figure~\ref{fig:failure_modes} (a), many grasps from different clusters are still concentrated in similar regions of the object, limiting the effectiveness of the diversity-based selection process.

\textbf{Ambiguous Query Images:} When the query image is captured from certain viewpoints, some objects or parts of objects in the image may appear ambiguous. For example, as illustrated in Figure~\ref{fig:failure_modes} (b), when tasking VLM to select a grasp that grips the lid of the given bottle, it is hard to tell from the image whether the candidate grasps are gripping the lid or the body. This ambiguity can lead to suboptimal grasp suggestions by the VLM.

\textbf{VLM Limitations:} The VLM itself has inherent constraints, including challenges in reasoning about complex spatial relationships and physical constraints in 3D scenes \cite{ramakrishnan2024does}. As shown in Figure~\ref{fig:failure_modes}(c), the VLM selects a grasp that results in instability, while in Figure~\ref{fig:failure_modes}(d), the contact point specified by the VLM is completely misaligned with the object.

\subsection{View‑Selection for Ambiguity Reduction}

\begin{figure}[tb]  
  \centering
  \includegraphics[width=\columnwidth]{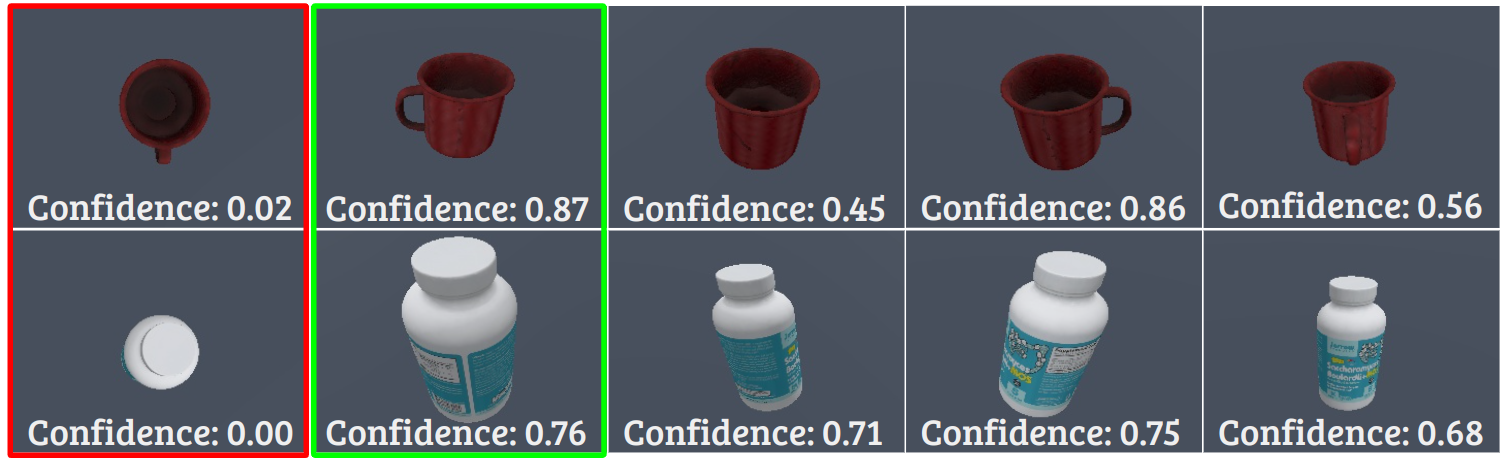}
  \caption{Illustration of using side cameras to obtain less ambiguous images. The first column (in red) shows the default hand-camera view, the second column (in green) shows the view with the highest confidence.}
  \label{fig:side_cams}
\end{figure}

\begin{table}[t]
\setlength{\belowcaptionskip}{-8pt}
\centering
\caption{Effect of view‑selection (Gemini + CPG).}
\label{tab:view_select}
\begin{tabular}{|l|c|c|c|}
\hline
\textbf{Input} & \textbf{Grasp} & \textbf{Task} & \textbf{Combined} \\ \hline
Default Hand camera (orig.) & 0.796 & 0.778 & 0.611 \\ \hline
Least‑ambiguous view & 0.778 & \textbf{0.833} & \textbf{0.685} \\
\hline
\end{tabular}
\end{table}

As discussed in Fig.~\ref{fig:failure_modes}(b), the single frame can obscure critical affordance cues (e.g., bottle neck vs.\ body), leading the VLM to confuse task semantics. We hypothesize that view ambiguity comes from the image itself, not the object. If another readily available viewpoint depicts the same scene with higher semantic clarity, the VLM should perform better.

To test that, we mount four static RGB cameras around the workspace, evenly spaced at 90° on a 0.2m radius circle. For each trial we:

\begin{enumerate}
  \item Capture multiple camera frames as described above
  \item Run Detic \cite{zhou2022detecting} with an text prompt matching the task object to obtain the maximum class–specific confidence $\mathrm{conf}_m$ in each image. (Illustrated in Fig.~\ref{fig:side_cams}).
  \item Select the image $m^\star=\arg\max_m \mathrm{conf}_m$ as least ambiguous and feed it to the standard TOG pipeline.
\end{enumerate}

Table~\ref{tab:view_select} compares results using the default hand-camera versus least-ambiguous view. While the Grasp Success Rate drops slightly, Task Compliance and Combined Success improve markedly (by 5.5 \% and 7.4 \%, respectively), notably improves task performance by reducing image ambiguity.

\section{Real-world Experiment}
\label{experiment2}


\begin{figure*}[t!]
  \centering
  \begin{tabular}{cccc}
    \includegraphics[width=0.224\textwidth, trim=0 120 0 230, clip]{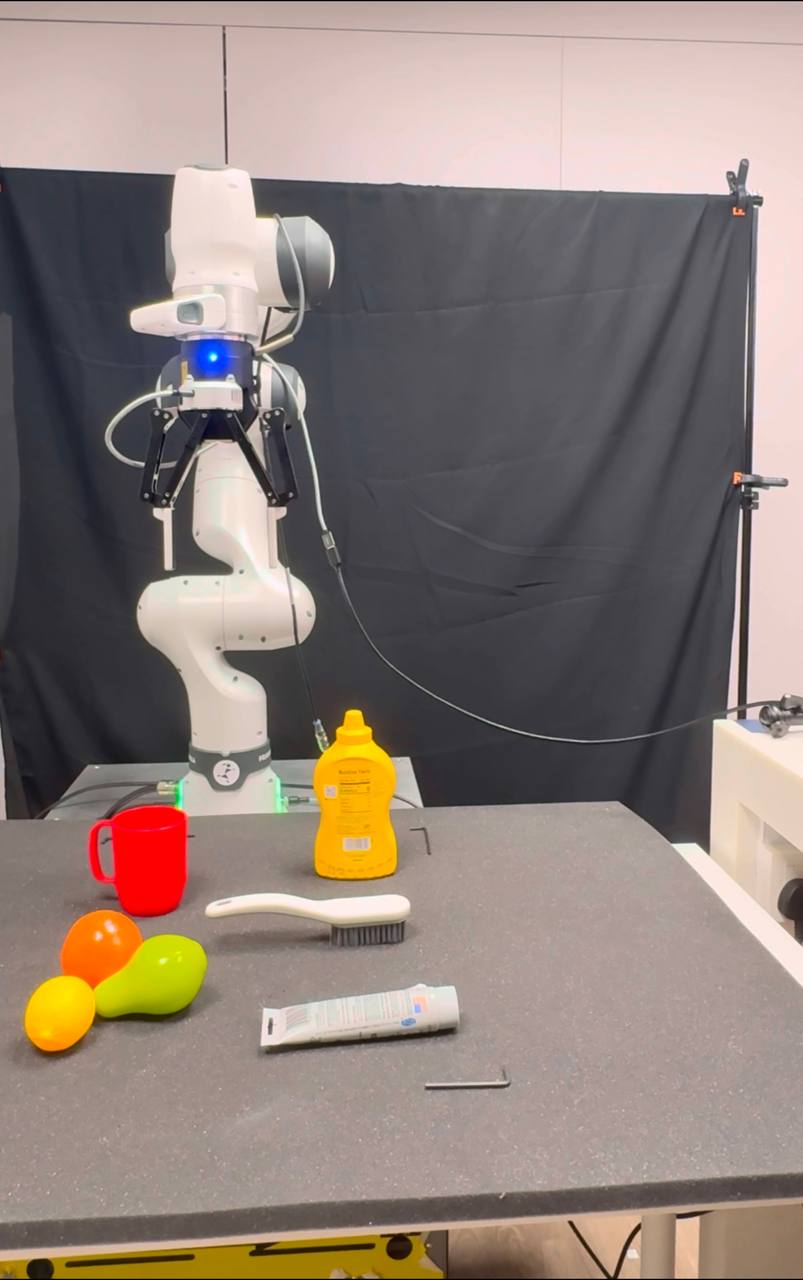} & 
    \includegraphics[width=0.224\textwidth, trim=0 120 0 230, clip]{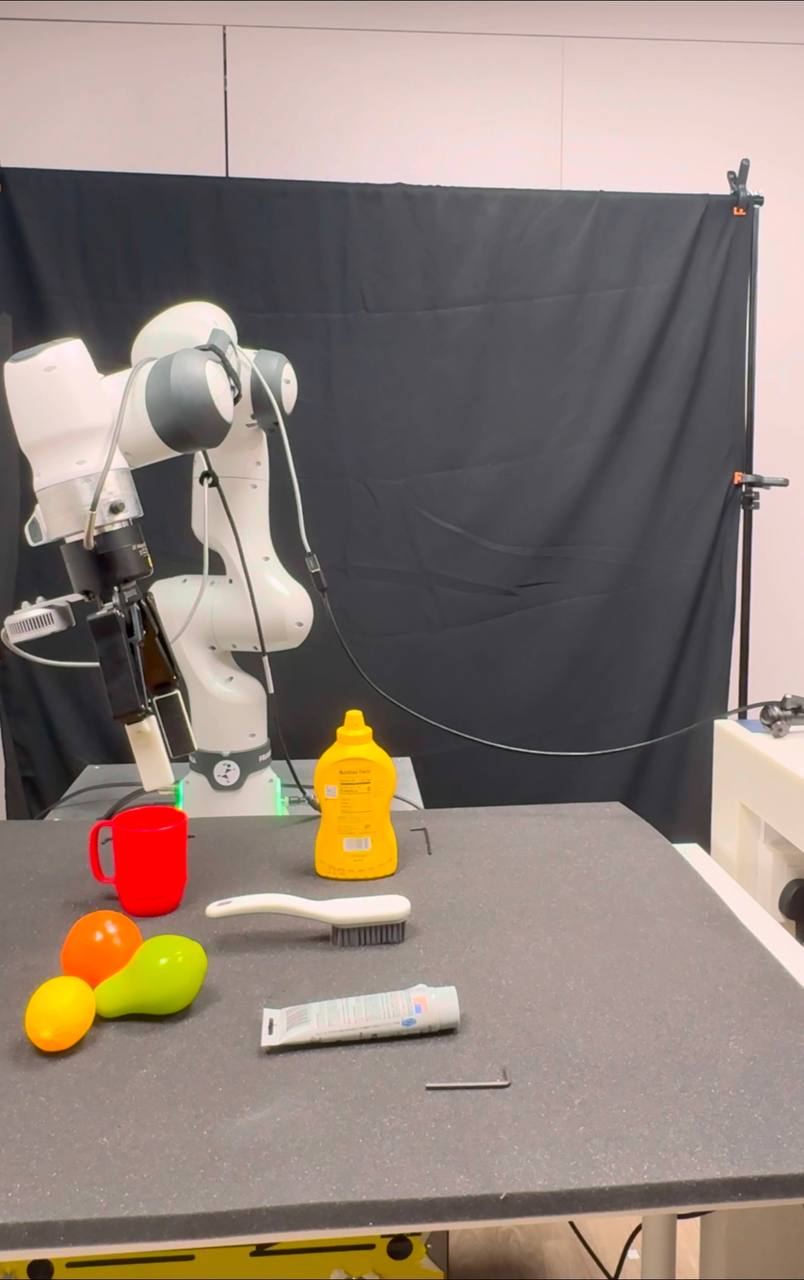} & 
    \includegraphics[width=0.224\textwidth, trim=0 120 0 230, clip]{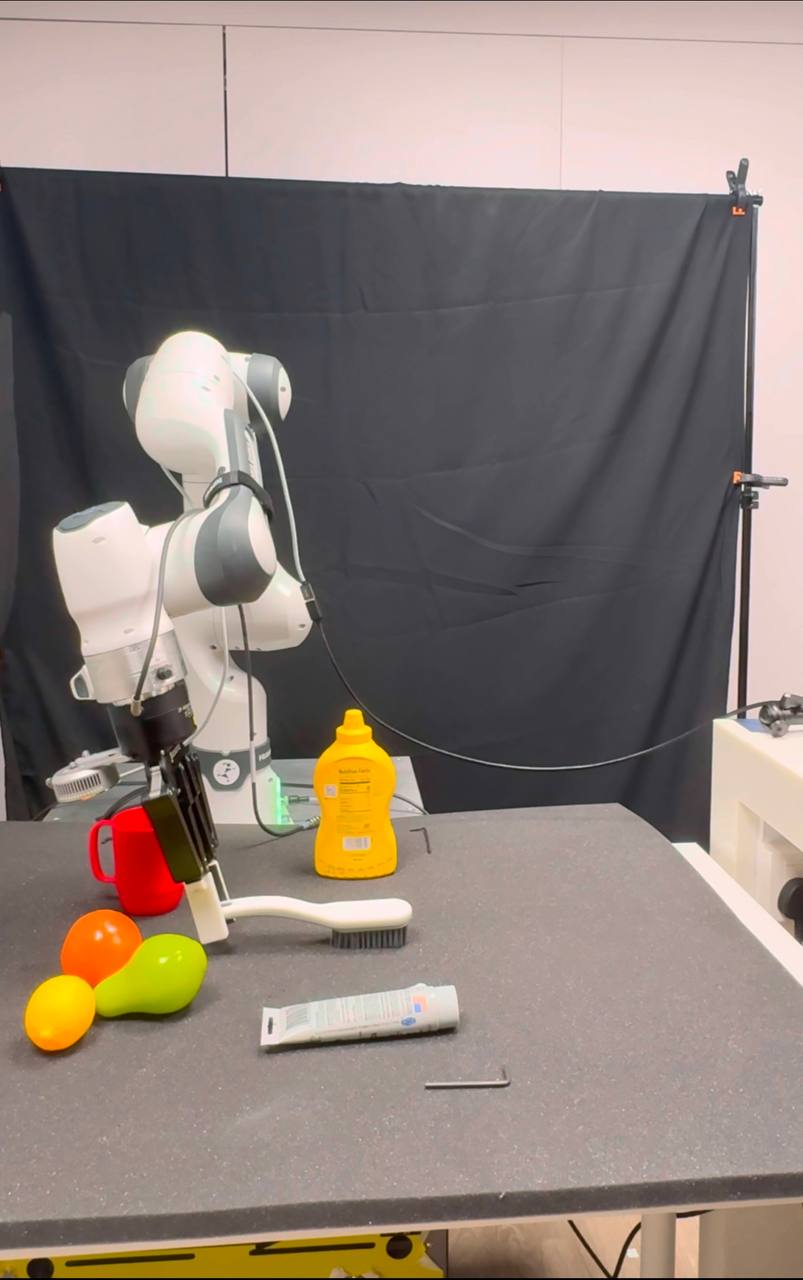} & 
    \includegraphics[width=0.224\textwidth, trim=0 120 0 230, clip]{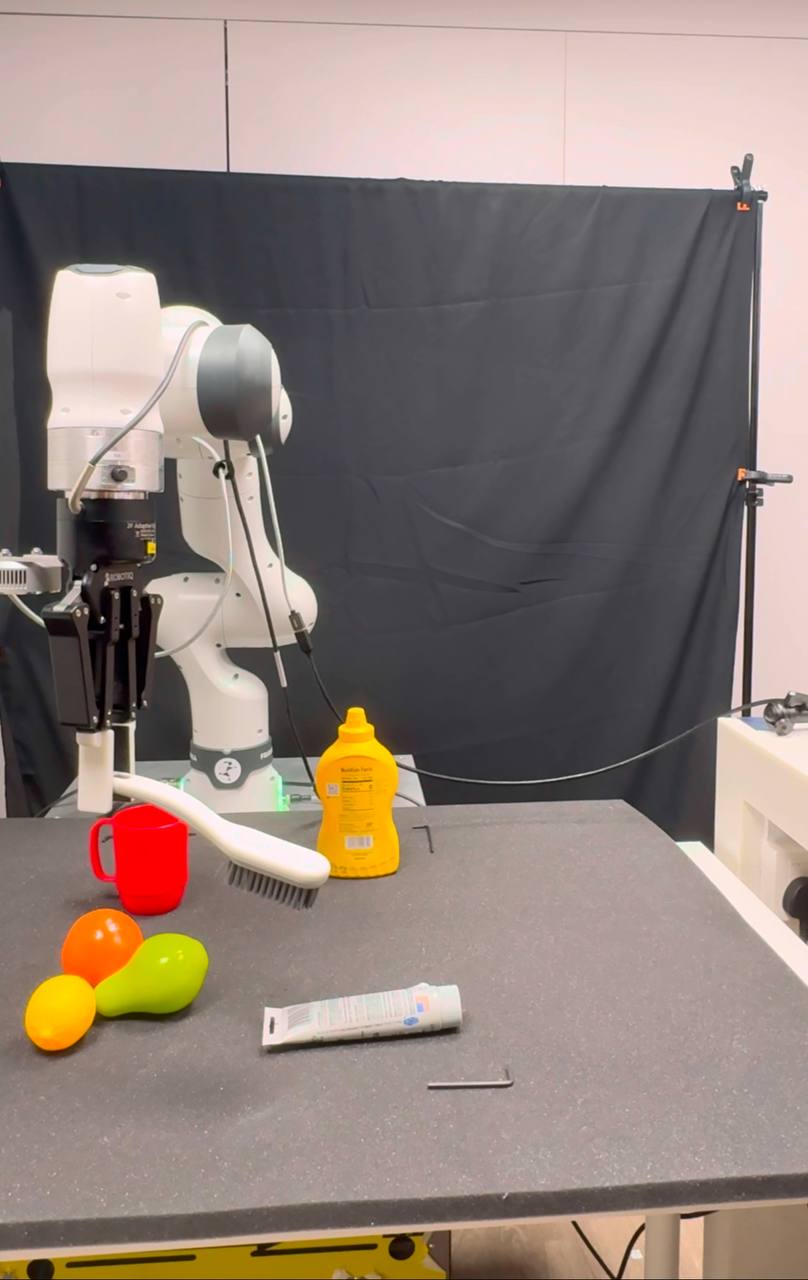} \\
    \includegraphics[width=0.224\textwidth, trim=0 120 0 230, clip]{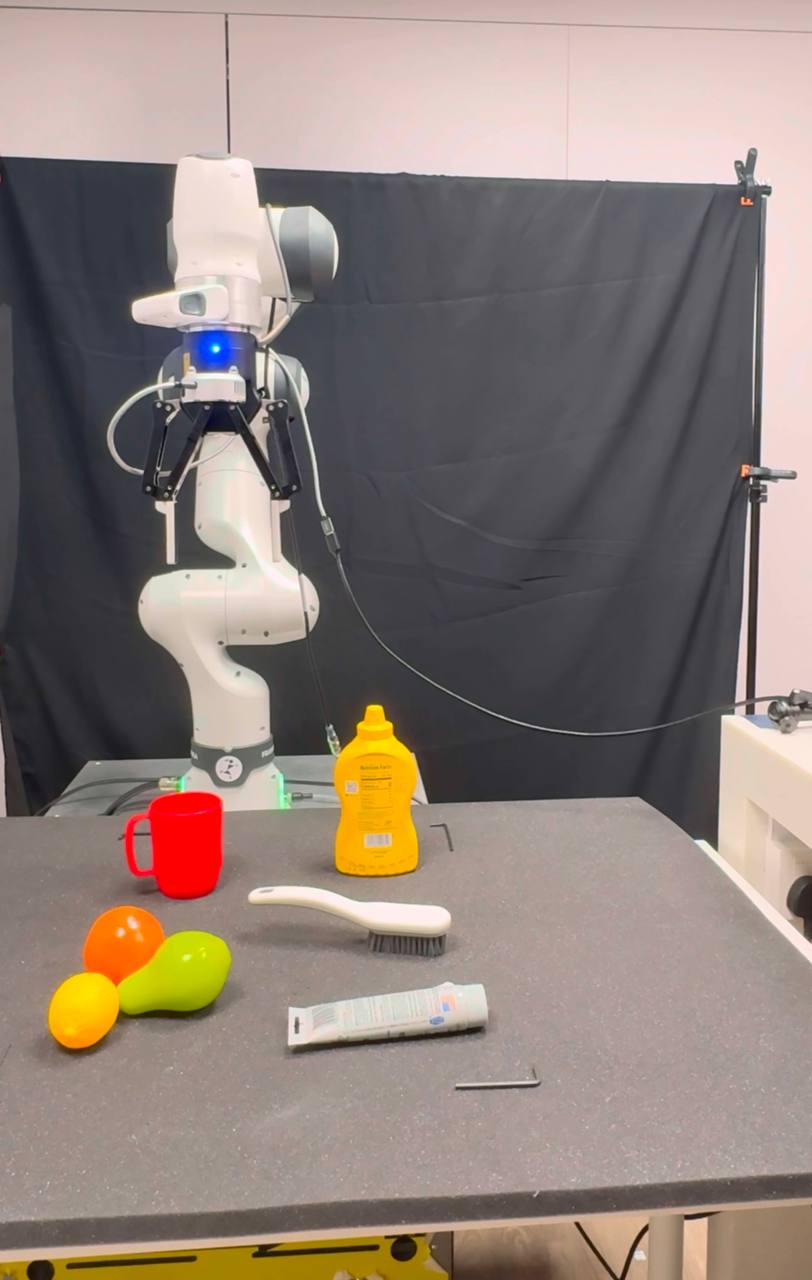} & 
    \includegraphics[width=0.224\textwidth, trim=0 120 0 230, clip]{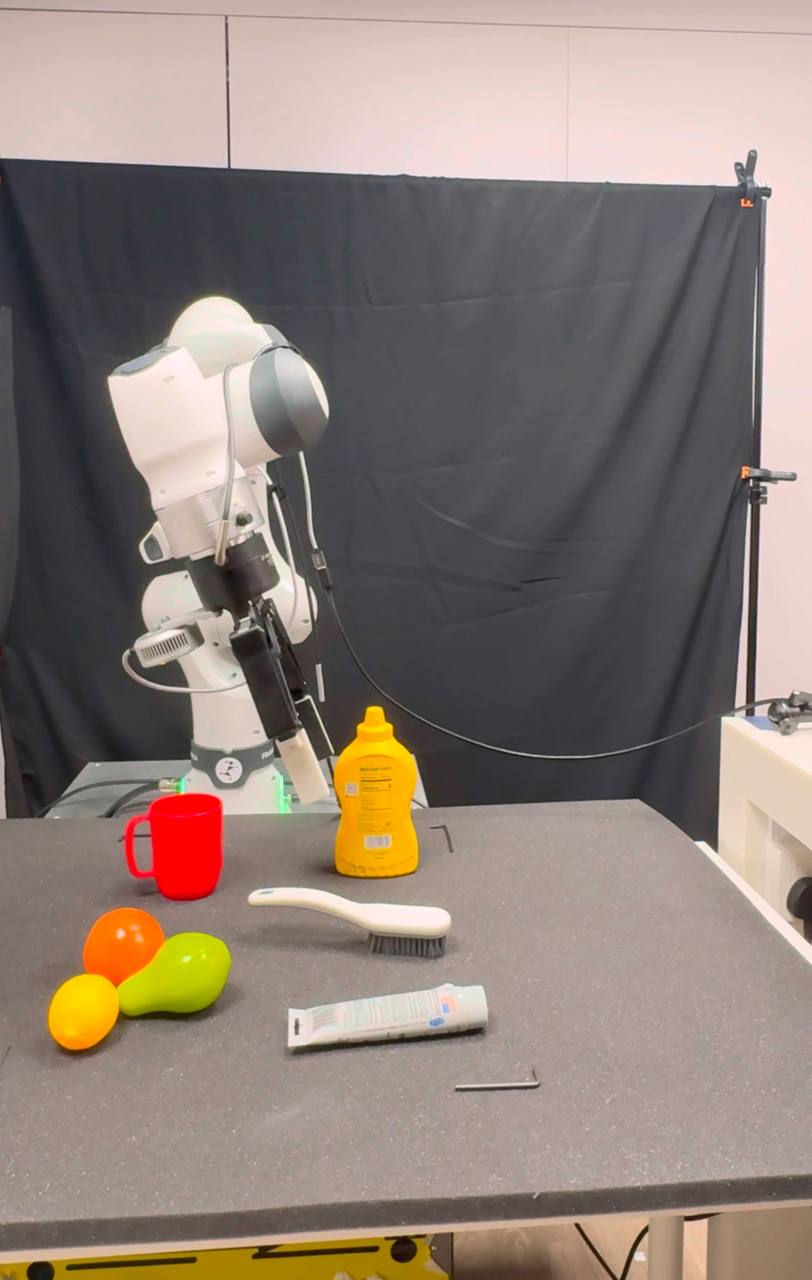} & 
    \includegraphics[width=0.224\textwidth, trim=0 120 0 230, clip]{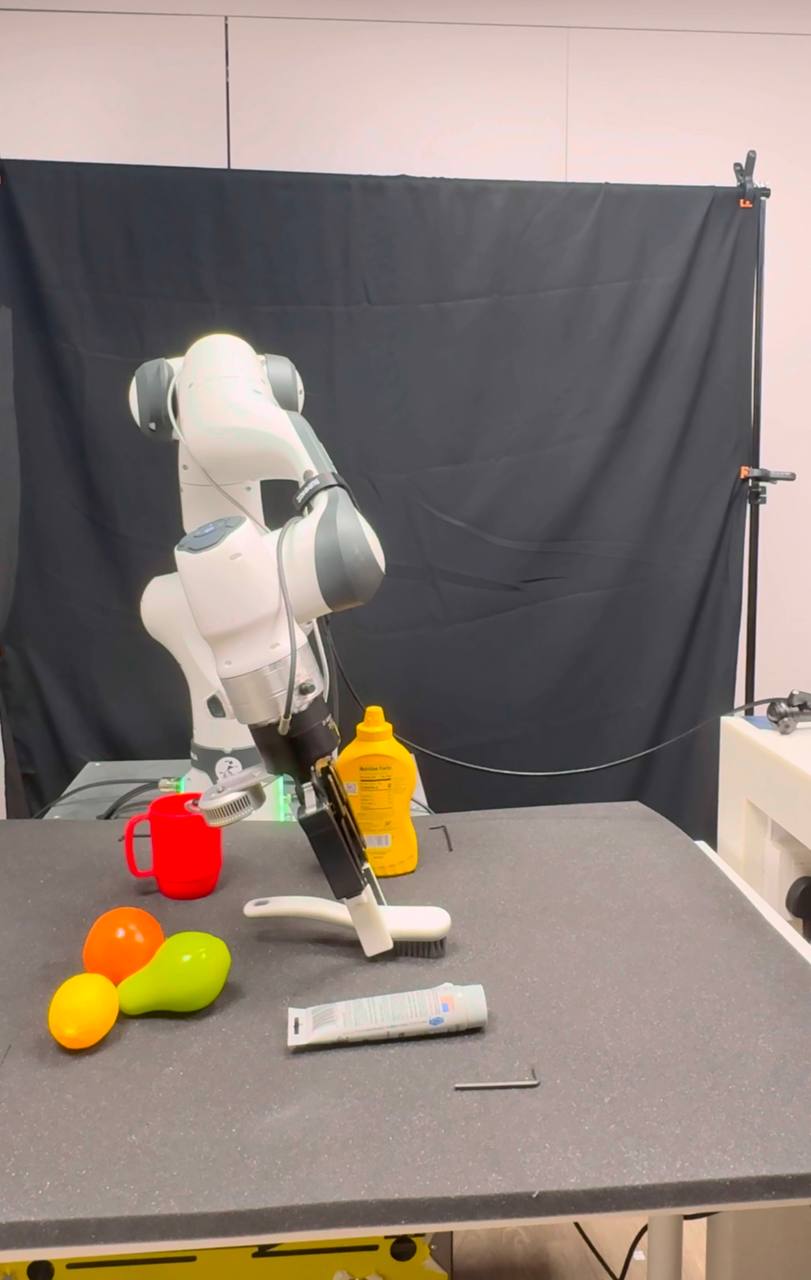} & 
    \includegraphics[width=0.224\textwidth, trim=0 120 0 230, clip]{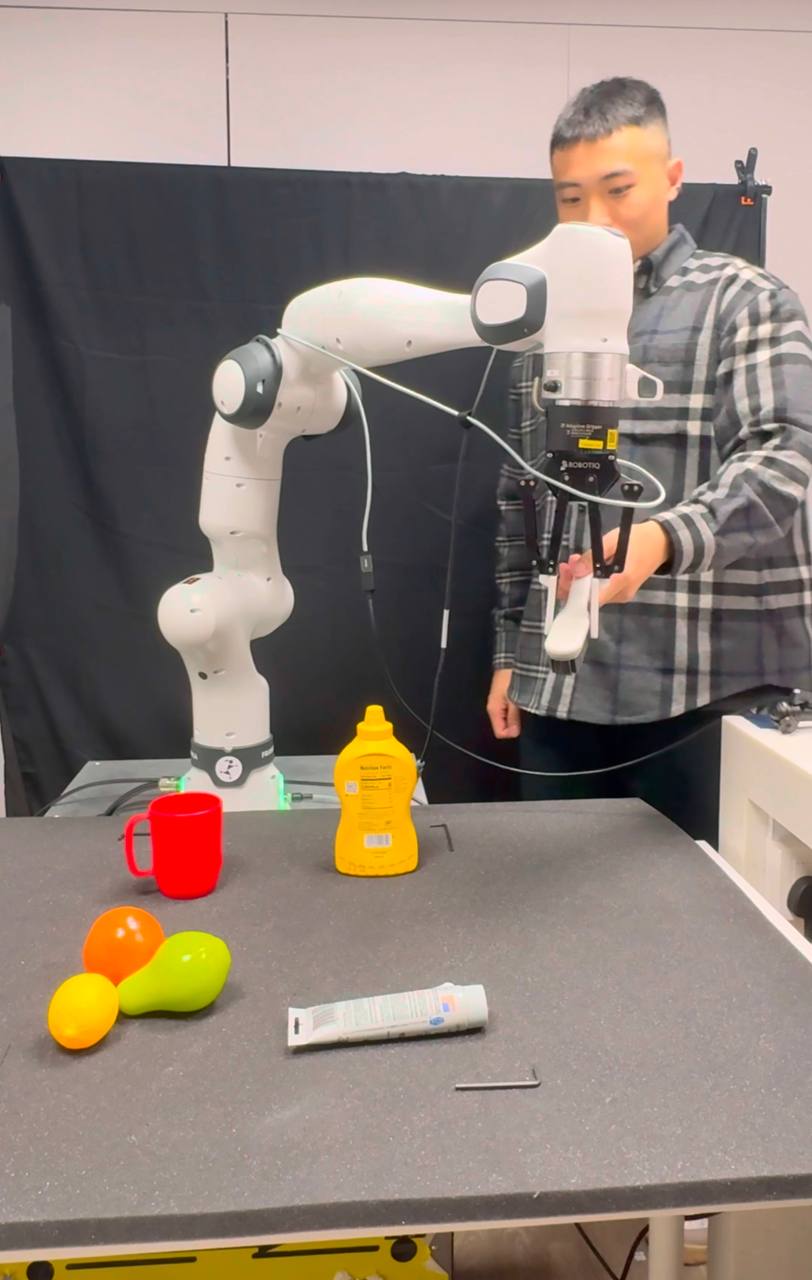} \\
  \end{tabular}
  \caption{The top row illustrates the task "use the brush," while the bottom row corresponds to the task "pass me the brush." The robot selects different grasping points based on the task description, demonstrating its ability to adapt grasp strategies according to task-oriented requirements.}
\end{figure*}

To test if our proposed method works in the real world, we extend the evaluation of our proposed method beyond simulation environments to assess its performance in real-world robotic manipulation tasks. 

\subsection{Experimental Setup}
Our hardware configuration employed a Franka Emika Panda robotic arm equipped with an Intel RealSense L515 RGB-D camera mounted on the end-effector link (eye-in-hand). We use the RGB and depth image to generate the point clouds. We use a single image to query the VLM and generate the grasps, thus mimic our setup in the simulator. 

We utilized  (CGN) \cite{sundermeyer2021contact} to generate unconditional grasps for each object and used Gemini-2.0-Pro~\cite{pichai2024gemini} as the VLM model. We use the CPG (Contact Point Generation) strategy as it performs the best~\ref{table:performance}, in which we directly ask the VLM to generate the best grasp point that satisfying the task description. Interesting, we find that the VLM can correctly select the corresponding grasps for the target object, without using an object detection and segmentation pipeline as in the common setup, indicating the use of VLM to further simplify the grasp generation pipeline.

\begin{figure}[h]
    \centering
    
    \begin{subfigure}{0.15\textwidth} 
        \includegraphics[width=\linewidth]{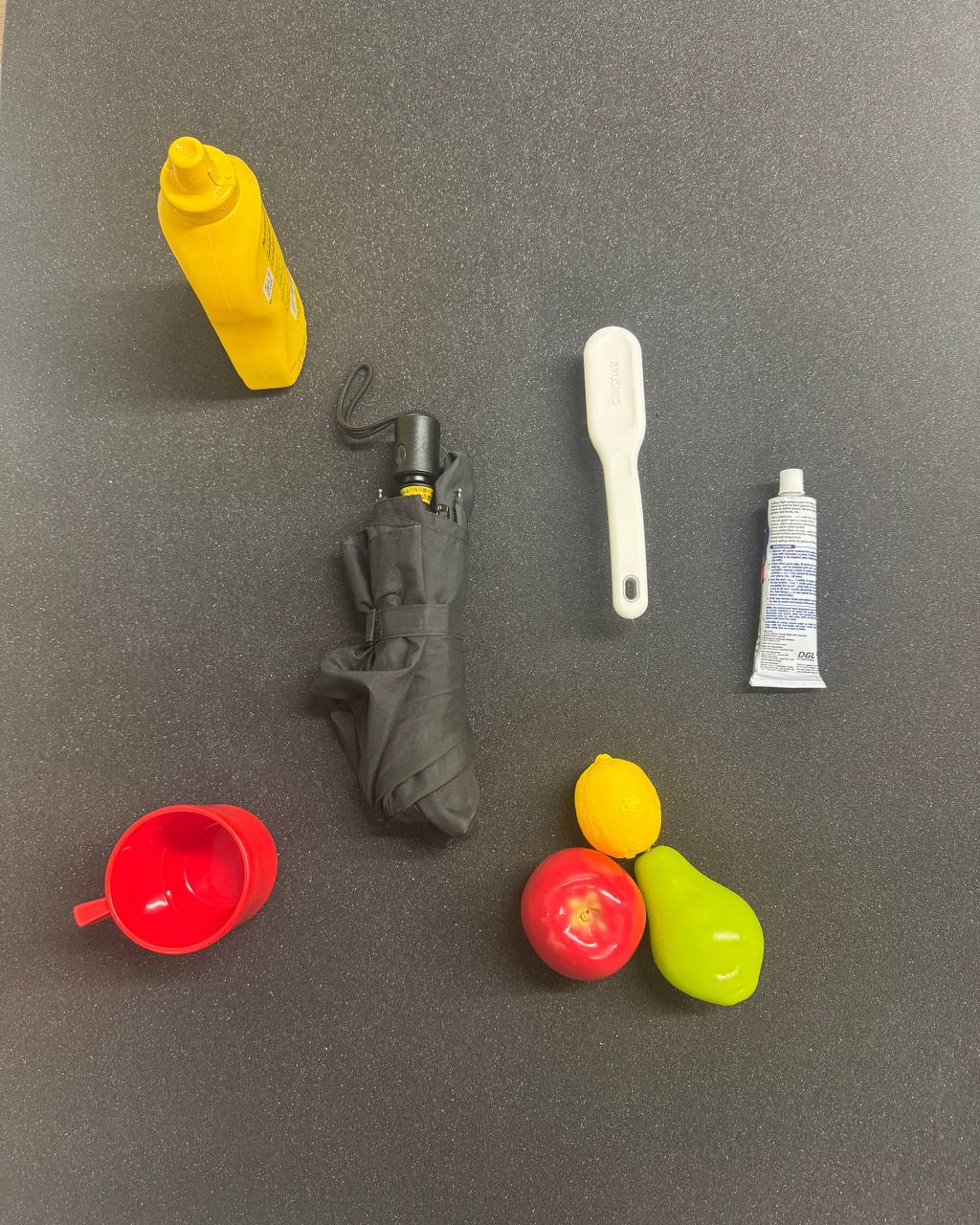}
        \caption{\small {Common Item}}
    \end{subfigure}
    \hfill
    \begin{subfigure}{0.15\textwidth}
        \includegraphics[width=\linewidth]{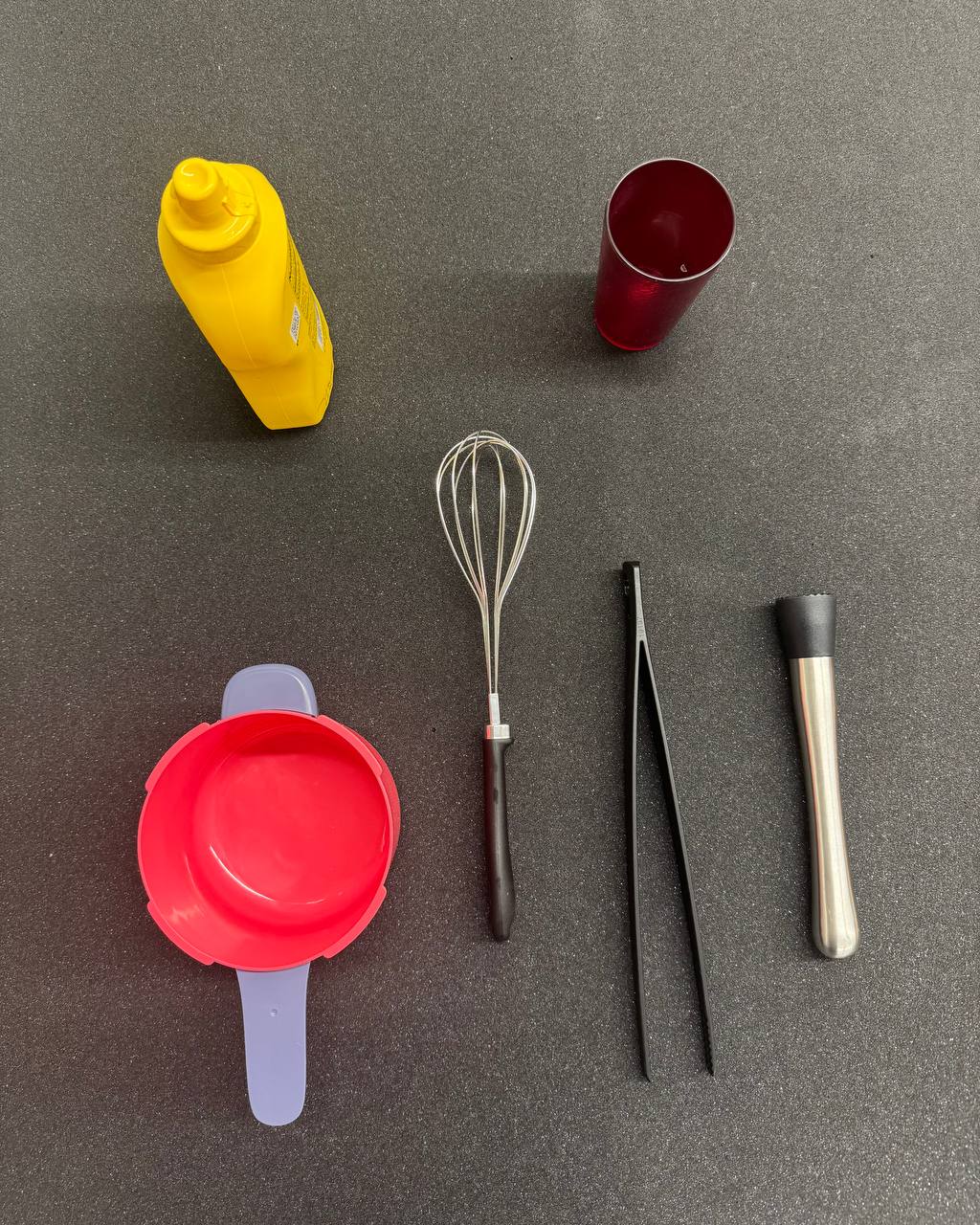}
        \caption{\small {Kitchen}}
    \end{subfigure}
    \hfill
    \begin{subfigure}{0.15\textwidth}
        \includegraphics[width=\linewidth]{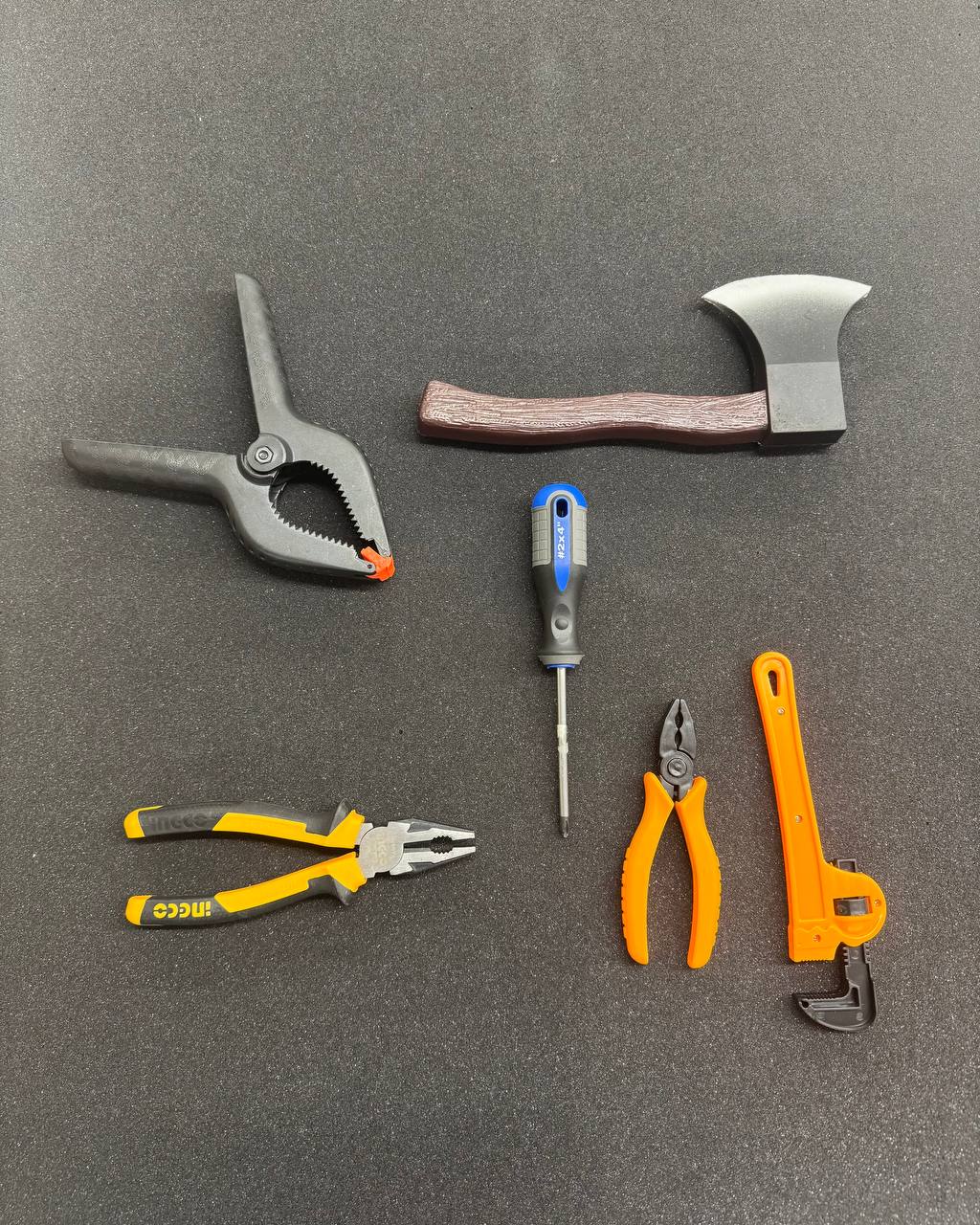}
        \caption{\small {Hand Tools}}
    \end{subfigure}
    \caption{Three real-world scenes}
    \label{fig:scenes}
\end{figure}

We evaluated grasp performance on 10 objects across three different scenes (Figure~\ref{fig:scenes}):

\begin{itemize}
    \item \textbf{Common Household Items:} The robot interacted with everyday objects, such as an umbrella, grey lotion, and a grey brush, arranged in cluttered settings.
    
    \item \textbf{Kitchen Utensils:} The robot handled various kitchen items, including a red pot, black spoon, red cup, whisk, and muddler, positioned among multiple objects.
    
    \item \textbf{Hand Tools:} The robot operated tools such as an axe and a screwdriver in densely arranged scenes.
\end{itemize}

For each object, we evaluated performance on two types of tasks:

\begin{itemize}
    \item \textbf{Use the Tool:} Success is measured by achieving a stable grasp appropriate for the intended use. For example, the robot should securely hold the red pot by its handle for cooking or firmly grip the whisk for mixing.
    
    \item \textbf{Handover:} Success is defined by passing the object to a person in a way that allows for easy and safe grasping. For instance, the robot should pass a hammer by grasping its head so that the user can easily take hold of the handle.
\end{itemize}

Each object-task combination was evaluated over three trials.

\begin{table}[htbp]
\setlength{\belowcaptionskip}{-10pt}
    \centering
       \caption{Real-world experiment results}
    \label{tab:real-world-experiment}
        \begin{tabular}{lccc}  
            \hline
            { \textbf{Metric}} & { \textbf{Household}} & { \textbf{Kitchen}} & { \textbf{Tools}} \\
            \hline
            { Grasp Success} & { 0.72} & { 0.90} & { 0.83} \\
            { Task Compliance} & { 0.94} & { 0.80} & { 0.83} \\
            { Combined Success} & { 0.68} & { 0.70} & { 0.75} \\
            \hline
        \end{tabular}

\end{table}

\subsection{Result}

Table~\ref{tab:real-world-experiment} presents the results of our real-world experiment. All three scenes consist of multiple objects, requiring the system to generate collision-free grasps. By integrating ContactNet's~\cite{sundermeyer2021contact} capability to generate collision-free grasps from partial point clouds with the VLM's ability to predict grasping points based on the task description, the TOG system remains both simple and effective. Requiring only a single pair of RGB and depth images as input, it achieves an overall success rate of 70\%.

We observe that the VLM generally achieves a high success rate in indicating reasonable grasp points. However, its performance declines when ambiguity arises due to the type of object and the viewpoint. For example, when viewed from a top-down perspective, the muddler's handle is difficult to distinguish, making it challenging to determine the correct grasping location. This aligns with our observations in the simulated experiment (Section~\ref{sec:failures}).

\section{Conclusion}
\label{conclusion}

In this paper, we introduced TOG, a training-free system for task-oriented grasp generation that integrates pre-trained grasp generation models with vision-language models (VLMs) to select grasps aligned with specific task requirements. We evaluated the proposed system in both simulated and real-world environments, demonstrating its effectiveness across various task setups. 

Crucially, we find that when leveraging a powerful VLM, simply prompting the model to indicate where to grasp and selecting from a set of feasible grasps generated by a pre-trained grasp generation model — the CPG strategy — yields the best performance. 

Despite its effectiveness, we identified several limitations, including insufficient grasp diversity, query image ambiguities, and constraints inherent to VLMs, highlighting areas for future improvement. Notably, we show that performance can be further improved through lightweight view selection to reduce perceptual ambiguity. This study underscores the potential of combining pre-trained grasp models with VLMs to enable flexible, efficient task-oriented manipulation without additional training, offering a promising framework for advancing robotic grasping capabilities.

\bibliographystyle{IEEEtran}
\bibliography{ref}

\end{document}